\documentclass[11 pt, conference]{ieeeconf}  
\IEEEoverridecommandlockouts       
\usepackage{geometry}
\usepackage{graphics} 
\usepackage{epsfig} 
\usepackage{mathptmx} 
\usepackage{times} 
\usepackage{amsmath} 
\usepackage{amssymb}  
\usepackage{multirow}
\usepackage{makecell}
\usepackage{tablefootnote}
\usepackage{algorithm} 
\usepackage{algorithmic} 
\usepackage{multirow} 

\begin{document}

\title{\LARGE \bf
Q-value Regularized Decision ConvFormer for Offline Reinforcement Learning}

\author{Teng Yan$^{1, *}$, Zhendong Ruan$^{1, *}$, Yaobang Cai$^2$, Yu Han$^2$, Wenxian Li$^2$, Yang Zhang${^2 \dag}$
\thanks{$^*$ Equal contribution, $^\dag$ Corresponding author.}%
\thanks{$^{1}$ College of Applied Sciences, Shenzhen University, China.}%
\thanks{$^{2}$ Sino-German College of Intelligent Manufacturing, Shenzhen Technology University, China.}%
\thanks{This research was completed while all the authors were working at the Mechanical Industry Key Laboratory of Intelligent Robotics Technology for 3C, Shenzhen Technology University.}
}

\maketitle 
\thispagestyle{empty}

\begin{abstract}
As a data-driven paradigm, offline reinforcement learning (Offline RL) has been formulated as sequence modeling, where the Decision Transformer (DT) has demonstrated exceptional capabilities. Unlike previous reinforcement learning methods that fit value functions or compute policy gradients, DT adjusts the autoregressive model based on the expected returns, past states, and actions, using a causally masked Transformer to output the optimal action. However, due to the inconsistency between the sampled returns within a single trajectory and the optimal returns across multiple trajectories, it is challenging to set an expected return to output the optimal action and stitch together suboptimal trajectories. Decision ConvFormer (DC) is easier to understand in the context of modeling RL trajectories within a Markov Decision Process compared to DT. We propose the Q-value Regularized Decision ConvFormer (QDC), which combines the understanding of RL trajectories by DC and incorporates a term that maximizes action values using dynamic programming methods during training. This ensures that the expected returns of the sampled actions are consistent with the optimal returns. QDC achieves excellent performance on the D4RL benchmark, outperforming or approaching the optimal level in all tested environments. It particularly demonstrates outstanding competitiveness in trajectory stitching capability. 
\end{abstract}

\section{Introduction}
Classic reinforcement learning (RL) is an online learning paradigm where an agent interacts with the environment iteratively to collect experiences and then uses those experiences to improve its policy to achieve maximum returns \cite{sutton1998introduction}. However, in many cases, online interactions are impractical due to the high cost or danger of data collection (e.g., in robotics, educational agents, or healthcare). To avoid such expensive and risky learning processes, offline reinforcement learning is adopted, which eliminates the need for direct environment interaction. This approach leverages pre-collected datasets to derive policies, making it a practical solution \cite{kumar2019data}.

Offline reinforcement learning has a paradigm more similar to supervised learning due to the fact that it learns from a dataset. A large portion of previously proposed offline reinforcement learning methods learn strategies based on constrained or regularized approximate dynamic programming (e.g., Q-learning or actor-critic methods) \cite{kumar2020conservative}.

DT\cite{chen2021decision} maximizes the likelihood of actions conditional on historical trajectories (including returns), which essentially transforms offline RL into supervised sequence modeling \cite{janner2021offline}.DT adjusts the autoregressive model to generate future actions based on the expected payoffs, past states, and action sequences as inputs to the information exchange through the attention module . More precisely, DT learns strategies conditional on RTG and states, and views the total of future rewards Return-to-go (RTG) as a goal. DT, which train by considering the future cumulative reward, or RTG, as the objective, encounter difficulty in achieving the integration capabilities necessary for offline RL agents. Specifically, the ability to combine segments of suboptimal trajectories to produce an optimal trajectory is a critical requirement \cite{kostrikov2021offline}.
DT's trajectory welding capabilities have been significantly improved by a significant amount of subsequent work. QDT employs dynamic programming to rename the RTG \cite{yamagata2023q}, thereby enhancing the DT by bringing the RTG closer to the true value. EDT optimizes trajectories by retaining lengthier histories when the previous trajectory is optimal and shorter histories when it is sub-optimal, allowing it to be "stitched" to a more optimal trajectory \cite{wu2024elastic}. The Reinformer\cite{zhuang2024reinformer} algorithm enhances splicing capabilities by adjusting the RTG to the utmost gain that can be achieved with the current historical trajectory. QT\cite{hu2024q} integrates the trajectory modeling capabilities of Transformer with the Dynamic Programming (DP) approach's predictability of optimal future payoffs to achieve exceptional performance in sequential reinforcement learning. This is accomplished by incorporating a Q-value module into the DT training process.

Nevertheless, reinforcement learning trajectories are distinct from conventional sequences, such as text or audio, and their modeling cannot be regarded as a straightforward sequence modeling task. Markov Decision Processes (MDPs) are the preferred method for defining reinforcement learning problems. In these processes, the probabilities of transitioning to the next state are determined solely by the current state and operation, rather than past states, as per the Markov property. Consequently, it is imperative to consider the local correlations between the stages of a trajectory sequence.The attention module of DT is over-parameterized and inadequate for capturing the unique local dependency patterns of MDPs. The Decision Convformer (DC), which utilizes a local convolutional filtering module, is more effective in capturing the local correlations that are intrinsic to RL datasets. Additionally, it is better adapted to capturing the local dependency patterns that are intrinsic to RL trajectories modeled as Markovian decision-making processes \cite{kim2023decision}.

The DC proposal significantly enhances DT; however, the splicing capability necessary for offline RL agents is not achieved when the sum RTG of future rewards is used as the training objective. Our newly modified framework in this paper enables conditional sequence reinforcement learning to more effectively capture the RL trajectories of Markov decision processes with outstanding trajectory splicing capability, as a result of the aforementioned understanding. The following are our contributions.

\begin{itemize}

\item We propose a method: Q-value Regularized ConvFormer a modified DC model and training method as an action learning predictor for offline reinforcement learning.
\item QDC exhibits exceptional average performance in the OpenAI Gym and exceptional splicing performance in the Maze2D task, resulting in a 10.57\% improvement over the strongest baseline. This enhances the splicing capabilities of the DC algorithm.
\item By utilizing higher-value actions and requiring fewer resources for training, QDC is capable of effectively capturing the local correlations that are inherent in RL datasets. This results in a faster training process.

\end{itemize}

\section{Preliminaries}

RL problems can be modeled as learning problems within a Markov Decision Process (MDP), \(M = \langle \rho_0, S, A, P, R, \gamma \rangle\), where \(\rho_0\) is the initial state distribution, \(S\) is the state space, \(A\) is the action space, \(P(s_{t+1}|s_t, a_t)\) is the transition probability, \(R(s_t, a_t)\) is the reward function, and \(\gamma \in (0,1)\) is the discount factor. The goal of traditional reinforcement learning is to find an optimal policy \(\pi^*\) that maximizes the expected return by interacting with the environment. Offline reinforcement learning, however, is conducted without interacting with the environment; it relies on an offline dataset \(D\) collected by a given policy \(D_\mu\). The objective of offline reinforcement learning is to learn a policy using the dataset \(D\) to maximize the expected return.

\subsection{Decision Transformer for Offline RL}
Transformers are among the most effective and scalable neural networks for modeling sequence data. Inspired by the tremendous success of sequence models in NLP, the Decision Transformer (DT) was proposed to model trajectory optimization problems as an action prediction process. DT transforms reinforcement learning (RL) problems into conditional sequence modeling tasks. In DT, trajectories are viewed as sequences of return-to-go (RTG), states, and actions. At each time step \(t\), DT uses a sub-trajectory of length \(K\) time steps as input:

{\footnotesize
\begin{equation}
\tau_t = (\hat{R}_{t-K+1}, s_{t-K+1}, a_{t-K+1}, \ldots,\hat{R}_{t-1}, s_{t-1}, a_{t-1}, \hat{R}_t, s_t)
\end{equation}
}

Using this \(\tau_t\), DT predicts the action \(a_t\). Here, \(\hat{R}_t\) denotes the return-to-go, defined as \(\hat{R}_t = \sum_{t' = t}^T r_{t'}\), 
representing the expected future rewards.

\begin{figure*}
\centering
\centerline{\includegraphics[width=\textwidth]{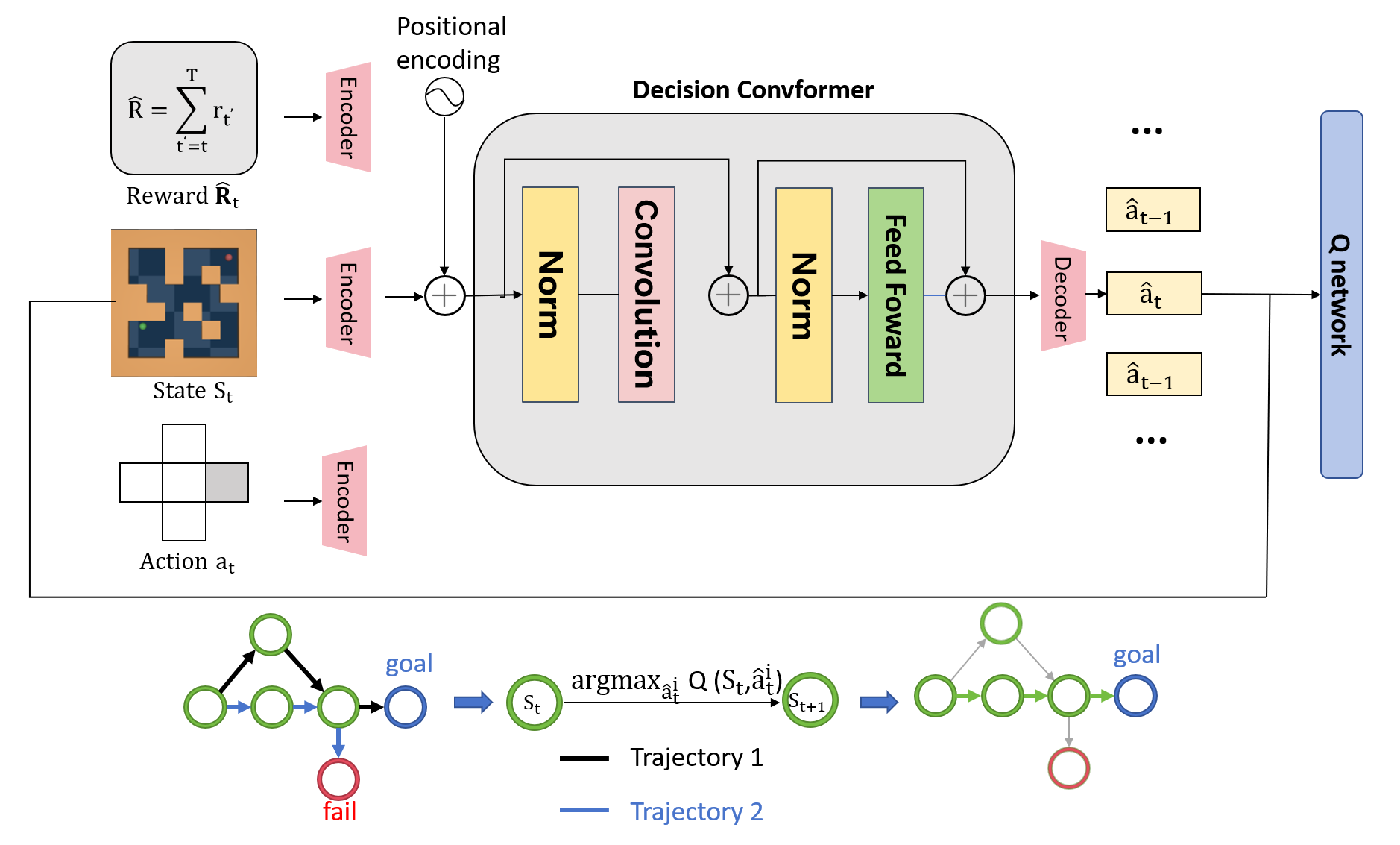}}
\caption{This is the QDC training framework. It uses state \(s\), action \(a\), and expected reward \(\hat{R}_t\) as inputs to predict the next action \(\hat{a}_t\). The predicted action \(\hat{a}_t\) and state \(s_t\) are then passed to the Q-network to learn the target network. Finally, QDC describes how, based on the Q-network, the policy is guided to take actions consistent with the expected return to stitch different trajectories into the optimal trajectory.}
\label{fig1}
\end{figure*}

\subsection{Decision ConvFormer for offline RL} 
Decision ConvFormer (DC) is an innovative action sequence predictor based on the MetaFormer architecture. DC employs local convolutional filters as an intermediate component, which effectively captures the inherent local dependencies in RL datasets. It processes multiple entities in parallel and understands the relationships between them through a general structure. Compared to DT, DC is better at capturing the dependencies in Markov Decision Processes within RL trajectories. It achieves state-of-the-art performance on offline reinforcement learning tasks while requiring fewer resources.

\subsection{Stitching in Conditional Sequence Modeling}
In offline reinforcement learning tasks, the challenge of stitching suboptimal trajectories into optimal ones has garnered significant attention. Theoretical work by Brandfonbrener et al. \cite{brandfonbrener2022does} and empirical findings by Kumar et al. \cite{kumar2022offline} have demonstrated that return-conditional sequence models lack the ability to perform stitching effectively. Both DT and DC treat offline reinforcement learning as a supervised learning task, where action prediction captures historical states and actions in the trajectories using causal models and convolutional models, respectively. QDT and EDT address trajectory stitching by reassigning RTG values to bias action predictions towards those with higher RTG. Reinformer, on the other hand, trains the model such that the predicted Rt corresponds to higher value actions. QDT incorporates a Q-learning module into the training process, where actions are predicted using a Transformer structure. The predicted actions and states are then used to train the Q-network, which in turn selects actions based on the Q-values.

\section{Q-value Regularized Convformer}

We propose a novel approach that combines the trajectory modeling capability of the Decision ConvFormer (DC) with the predictability of optimal future returns in dynamic programming (DP) methods, thereby creating an algorithm suited for offline reinforcement learning (RL) problems.
First, we detail the application of the conditional transformer policy as a behavior cloning expression strategy framework. Next, we describe how to incorporate the Q-value module into our transformer policy during the training phase and use the behavior cloning term as a regularization mechanism for the policy. Finally, we explain how to utilize the learned Q-value function for inference.

\subsection{Model Architecture} 
Our model framework, illustrated in Figure~\ref{fig1}, replaces the Decision Transformer's causal transformer with a convolution module, transforming it into the Decision ConvFormer model. The DC model inputs include multiple states \(s_t\), actions \(a_t\), and return-to-go \(R_t\), which output expected future sequences from historical data. During training with offline data, the Decision Transformer (DT) processes the trajectory sequence \(\tau_t\) autoregressively, incorporating the last \(K\) steps of historical context.

The DT model trains on input sequences of state labels \(s_t\), return-to-go \(R_t\), and their corresponding actions \(a_t\) to predict the subsequent actions \(a_t\). Each state \(s_t\) and return-to-go \(R_t\) are used as supervisory signals to output \(a_t\).

For continuous action spaces, the training objective minimizes the mean square error:

\begin{equation}
L_{DC} = \mathbb{E}_{\tau_t \sim D} \left[ \frac{1}{K} \sum_{i=t-K+1}^t \left(a_i - \pi(\tau_t)_i\right)^2 \right]
\end{equation}

Our approach employs the same input method. The actions \(a_t\) and corresponding states \(s_t\) are fed into the DC model for Q-value training. Finally, the learned target Q-value function is used to infer actions for the encountered states \(s_t\).

\subsection{Convolution Module} 
In DC, for each time step \(t\), the input sequence is in the form:

\begin{equation}
\tau_t = (\hat{R}_{t-K+1}, s_{t-K+1}, a_{t-K+1}, \ldots, \hat{R}_t, s_t, a_t)
\end{equation}

Each RTG, state, and action \(a_t\) is embedded individually, producing:

\begin{equation}
\tau_t = \left[\begin{array}{c}
\text{Emb}_{\hat{R}} (\hat{R}_{t-K+1}); \\
\text{Emb}_s (s_{t-K+1}); \\
\text{Emb}_a (a_{t-K+1}); \\
\vdots \\
\text{Emb}_{\hat{R}} (\hat{R}_t); \\
\text{Emb}_s (s_t)
\end{array}\right] \in \mathbb{R}^{(3K-1) \times d}
\end{equation}

where the sequence length is \(3K-1\) and \(d\) denotes the hidden dimension. The sequence \(\tau_t\) then passes through \(N\) stacked convolutional blocks, each containing two sub-blocks.

The first sub-block involves layer normalization followed by a convolution module for token mixing:

\begin{equation}
  Z_t^{\text{1st sub-block}} = \text{Conv}(\text{LN}(\tau_t)) + \tau_t
\end{equation}

The second sub-block involves layer normalization followed by a feedforward network:

{\footnotesize
\begin{equation}
Z_t^{\text{2nd sub-block}} = \text{FFN}(\text{LN}(Z_t^{\text{1st sub-block}})) + Z_t^{\text{1st sub-block}}
\end{equation}
}

The FFN is implemented as a two-layer MLP. Finally, the action \(a_t\) is output.

Compared to DT, DC has significantly fewer parameters than the attention modules of DT. Additionally, the convolution modules can integrate temporal information between adjacent tokens. The convolution module in DC is well-suited for capturing the inherent local dependency patterns in RL trajectories modeled as Markov decision processes. This enhances the understanding of the underlying data, improving the overall performance of the model. DC shows significant advantages and broad application prospects in handling complex sequential data and achieving intelligent decision-making.

\subsection{Inference with Q-value Module} 
In both DT and DC algorithms, the inputs are recalculated and labeled RTGs. Since RTG represents the expected future rewards, which are challenging to predict accurately, it can affect the generation of actions \(a_t\) within the sequence. Solely using RTG for outputting action trajectories may not yield optimal action sequences. To address the challenge of stitching together trajectories and formulating a strategy that aligns the expected returns of sampled operations with the optimal returns, we employ a Q-value module training framework.

Specifically, our training framework follows the approach of QT, combining the trajectory modeling capability with the predictability of optimal future returns from DP methods. DC serves as the action predictor to generate actions, while the learned Q-network selects high-reward actions and refines DC.

We learn the Q-value function in the traditional manner by minimizing the Bellman\cite{fujimoto2019off} operator's target network \(Q_{\phi_1}\) and \(Q_{\phi_2}\), using the double Q-learning technique. We construct two Q-networks \(Q_1\) and \(Q_2\) and their respective target networks \(Q_{\phi_1'}\), \(Q_{\phi_2'}\), and policy \(\pi_{\theta'}\). Given that the transformer policy's input includes trajectory history, we use an \(n\)-step Bellman equation to estimate the Q-value function. This is achieved by minimizing the following equation:

\begin{equation}
\mathbb{E}_{\tau_t \sim D, \hat{a}_t \sim \pi_{\theta'}} \sum_{m=t-K+1}^{t-1} \left| \left| \hat{Q}_m - Q_{\phi_i}(s_m, a_m) \right| \right|^2
\end{equation}

where

\begin{equation}
\hat{Q}_m = \sum_{j=m}^{t-1} \gamma^{j-m} r_j + \gamma^{t-m} \min_{i=1,2} Q_{\phi_i'}(s_t, \hat{a}_t)
\end{equation}

and \(\phi_i\) for \(i \in \{1, 2\}\).

Here, \(\gamma\) is the discount factor, and \(\hat{a}_t\) denotes the predicted action output by the target model \(\pi_{\theta'}\).

By integrating the Q-value module during the training phase, we prioritize sampling high-value actions. The ultimate policy learning objective is a linear combination of policy regularization and policy improvement terms:

{\footnotesize
\begin{align}
\pi &= \arg \min_{\pi_\theta} \left\{ L(\theta) := L_{DC}(\theta) + L_Q(\theta) \right\} \notag \\
    &= \arg \min_{\pi_\theta} \left\{ L_{DC}(\theta) - \alpha \cdot \mathbb{E}_{\tau_t \sim D} \mathbb{E}_{(s_i, a_i) \sim \tau_t} Q_\phi(s_i, \pi(\tau_t)_i) \right\}
\end{align} 
}

Considering the scale variations of Q-value functions across different offline datasets, we adopt the normalization technique by Fujimoto \& Gu\cite{fujimoto2021minimalist}. \(\alpha\) is defined as
\begin{equation}
\alpha = \frac{\eta}{\mathbb{E}_{\tau_t \sim D} \mathbb{E}_{(s, a) \sim \tau_t} \left[ |Q_\phi(s, a)| \right]}
\end{equation}

Finally, during inference, the Q-value module samples multiple candidate RTG tokens and outputs actions based on different returns, selecting the action with the highest Q-value given \(s_t\). This Q-value regularization approach enhances the policy by prioritizing high-value behaviors, closely aligning the learning process with the optimal returns.
\begin{table*}[htbp]
\centering
\caption{The performance of QDC and SOTA baselines on D4RL Gym, Adroit and Maze2D tasks}\label{tab1}
\begin{tabular}{|c|c|c|c|c|c|c|c|c|c|c|}
\hline\hline
\multicolumn{2}{|c}{\textbf{Dataset}} & \multicolumn{3}{c}{\textbf{Value-Based Method}} & \multicolumn{6}{c|}{\textbf{Return-Conditioned BC}} \\
\cline{1-11}
\multicolumn{2}{|c|}{} & \textbf{BC} & \textbf{CQL} & \textbf{IQL} & \textbf{DT} & \textbf{QDT} & \textbf{Reinformer} & \textbf{DC} & \textbf{QT} & \textbf{QDC (OURS)} \\
\hline
\multirow{10}{*}{\makecell{\textbf{Gym}\\\textbf{Tasks}}} 
& halfcheetah-medium & 75.3 & 49.2 & 47.4 & 42.6 & 39.3 & 42.94 & 43 & 51.4 & 47.4±0.9 \\
& halfcheetah-medium-replay & 36.6 & 45.5 & 44.2 & 36.6 & 35.6 & 39.01 & 41.3 & 48.9 & 44.4±0.2 \\
& halfcheetah-medium-expert & 55.2 & 91.6 & 86.7 & 86.8 & - & 92.04 & 93 & 96.1 & 90.8±4 \\
& hopper-medium & 42.6 & 69.4 & 66.3 & 67.6 & 66.5 & 80.02 & 92.5 & 96.9 & 92.8±0.9 \\
& hopper-medium-replay & 18.1 & 95 & 94.7 & 82.7 & 52.1 & 83.67 & 94.2 & 102 & 99.6±0.6 \\
& hopper-medium-expert & 52.5 & 105.4 & 91.5 & 107.6 & 67.1 & 107.82 & 110.4 & 113.4 & 112.9±0.5 \\
& walker2d-medium & 52.9 & 83 & 78.3 & 74 & - & 80.52 & 79.2 & 88.8 & 86.8±0.3 \\
& walker2d-medium-replay & 32.3 & 77.2 & 73.9 & 79.4 & 58.2 & 72.89 & 76.6 & 98.5 & \textbf{99.4}±0.6 \\
& walker2d-medium-expert & 107.5 & 108.8 & 109.6 & 108.1 & - & 109.35 & 109.6 & 112.6 & 112.3±1.2 \\
\cline{2-11}
& \textbf{Average} & 52.6 & 80.6 & 77 & 76.2 & - & 78.7 & 82.2 & 89.8 & 87.4 \\
\hline
\multirow{4}{*}{\makecell{\textbf{Maze2D}\\\textbf{Tasks}}}
& maze2d-umaze & 88.9 & 94.7 & 42.1 & 31 & 57.3 & 57.15 & - & 105.4 & \textbf{160.8}±9.1 \\
& maze2d-medium & 38.3 & 41.8 & 34.9 & 8.2 & 13.3 & 85.62 & - & 172 & \textbf{172.8}±1.7 \\
& maze2d-large & 1.5 & 49.6 & 61.7 & 2.3 & 31 & 47.35 & - & 240.1 & 238.5±7.3 \\
\cline{2-11}  
& \textbf{Average} & 42.9 & 62.0 & 46.2 & 13.8 & 33.8 & 63.4 & - & 172.5 & \textbf{190.7} 
\\
\hline
\multirow{7}{*}{\makecell{\textbf{Adroit}\\\textbf{Tasks}}}
& pen-human-v1 & 63.9 & 37.5 & 71.5 & 79.5 & - & - & - & 129.6 & 124.8±3.2 \\
& hammer-human-v1 & 1.2 & 4.4 & 1.4 & 3.7 & - & - & - & 35.6 & 32.2±1.4 \\
& door-human-v1 & 2.0 & 9.9 & 4.3 & 14.8 & - & - & - & 28.7 & \textbf{30.95}±2.3 \\
& pen-cloned-v1 & 37.0 & 39.2 & 37.3 & 75.8 & - & - & - & 125 & 114.5±4.2 \\
& hammer-cloned-v1 & 0.6 & 2.1 & 2.1 & 3 & - & - & - & 23 & 19.4±2.8 \\
& door-cloned-v1 & 0.0 & 0.4 & 1.6 & 16.3 & - & - & - & 20.6 & 19.5±1.2 \\
\cline{2-11}
& \textbf{Average} & 17.5 & 15.6 & 19.7 & 32.2 & - & - & - & 60.4 & 56.9 \\
\hline
\end{tabular}
\end{table*}

\section{Experiments}

In this section, we conduct a comprehensive evaluation of our QDC model using the D4RL benchmark \cite{fu2020d4rl}, which includes not only the Gym datasets with near-saturated performance but also the challenging Maze2D and Adroit datasets. The main objectives of our experiments are:

\begin{itemize}
\item To compare the performance of the QDC model with other state-of-the-art (SOTA) benchmarks in offline RL.
\item To determine whether combining the DC model with Q-learning effectively adapts and improves its performance.
\item To evaluate the stitching ability of QDC for sub-optimal trajectories.
\item To assess the impact of the parameters in QDC on overall performance.
\end{itemize}

\subsection{Baselines}
We compare QDC with existing SOTA offline RL methods, including three value-based methods: Behavior Cloning (BC) \cite{pomerleau1988alvinn}, Conservative Q-Learning (CQL), and Implicit Q-Learning (IQL), as well as sequence modeling-based offline RL methods: Decision Transformer (DT), Q-learning Decision Transformer (QDT), Reinformer, Decision ConvFormer (DC), and Q-value Regularized Transformer (QT).

\section{Results}
\subsection{Main Results}
QDC is compared with baselines in three task domains, and the results are reported in Table~\ref{tab1}. Results for QDC correspond to the mean and standard errors of normalized scores over 30 random rollouts (3 independently trained models and 10 trajectories per model) for all tasks, which generally exhibit low variance in performance. Our method outperforms and is competitive with all existing methods across almost all domains, with particularly strong performance in the Maze2D domains. To ensure a fair comparison, scores are normalized according to the protocol established by Fu et al. \cite{yamagata2023q}, where a score of 100 corresponds to the expert policy. The analysis is conducted for each specific domain.

\subsubsection{Results for Gym Domain}
In the Gym tasks, both classical offline reinforcement learning algorithms and sequence modeling algorithms demonstrate proficiency. QDC, with its advantage of sequence modeling over traditional methods, achieves further improvements compared to sequence modeling algorithms like DT and DC, particularly exceeding traditional offline reinforcement learning methods and transformer-based methods in the 'medium' and 'medium-replay' tasks. Compared to the DC algorithm, QDC effectively enhances performance in Gym's 'medium', 'medium-replay', and 'medium-expert' tasks by combining Q-learning. Additionally, convolution-based QDC approaches or even surpasses the optimal performance of transformer-based QT across all Gym tasks.

\subsubsection{Results for Maze2D Domain}
Maze2D is an environment where forces drive a ball to move to a fixed target position and serves as a benchmark for evaluating the effectiveness of offline RL algorithms in stitching together different trajectory segments. QDC performs excellently in Maze2D, improving by 10.5\% compared to the strongest baseline, QT. The convolution module as an action predictor better captures local dependency patterns in RL trajectories, enhancing the ability to stitch suboptimal trajectories effectively.

\subsubsection{Results for Adroit Domain}
In the Adroit domain, due to the limited range of human demonstrations, offline reinforcement learning faces challenges from extrapolation errors, making strong policy normalization crucial. QDC, combined with Q-learning, exhibits excellent performance.

\subsection{Ablation Study}
Based on QDC's outstanding performance on D4RL tasks, we conduct ablation experiments to investigate the impact of QDC's main components on its overall effectiveness. We choose DC and QT as comparison baselines to highlight the differences between DC, QT, and our QDC.

\subsubsection{Stitching Ability}
The Maze2D domain, which involves a navigation task with a fixed target position, serves as a benchmark for evaluating the stitching capability of offline reinforcement learning algorithms in combining suboptimal trajectories into optimal ones. It includes four mazes of increasing difficulty—open, umaze, medium, and large—and uses two reward functions: normal and dense. Normal rewards are granted only when the target is achieved, while dense rewards are incrementally assigned at each step, inversely proportional to the distance from the target. Table~\ref{tab2} summarizes the results, presenting the average and standard deviation scores reported over 3 seeds for the Maze2D tasks. This table encompasses four increasingly complex mazes—open, umaze, medium, and large—each with two reward functions: normal and dense. The highest average scores are highlighted in bold.

\subsubsection{Long Task Horizon Ability}
\begin{figure}[htbp]
\centering
\centerline{\includegraphics[width=0.5\textwidth]{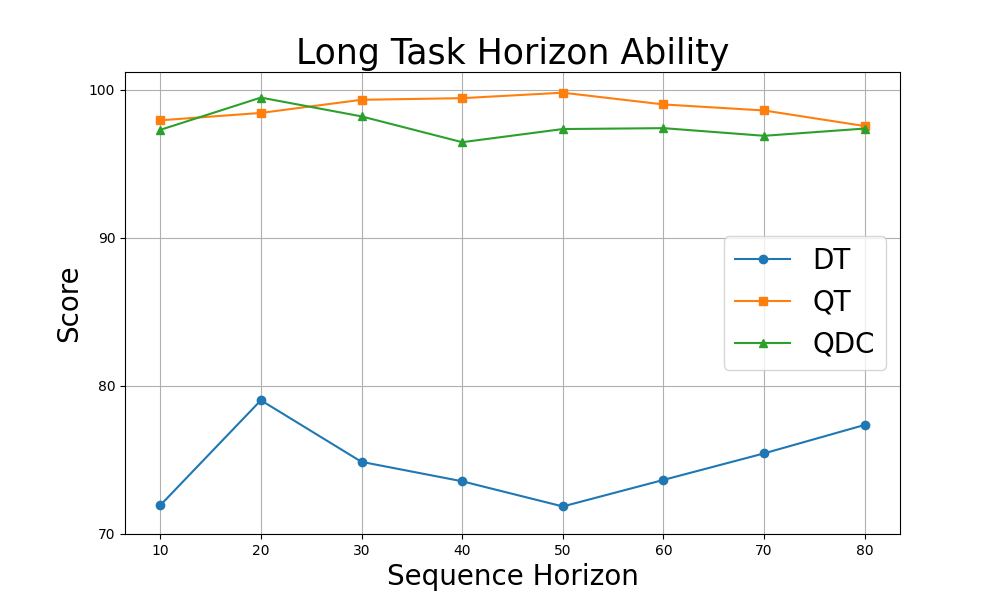}}
\caption{Ablation on the long task horizon ability. This encompasses the performance comparison of different input sequence horizons \(K \in [10, 80]\) in the walker2d-medium-replay-v2 task.}
\label{fig2}
\end{figure}

In Markov environments, the previous state is usually sufficient to determine the current action. However, DT experiments show that past information can be valuable for sequence modeling methods in certain environments, where longer sequences often yield better results (see Figure~\ref{fig2}). DC, as a predictor, is more suitable for capturing local dependencies in Markov RL datasets. We explore the impact of different sequence lengths on performance. We use the walker2d-medium-replay-v2 task to set different K input lengths and compare the effects on QT and QDC as the sequence length K varies. DT initially deteriorates after K=20 but recovers at K=80. QDC maintains stable performance compared to QT, demonstrating superior capability in managing extended task horizons.

\begin{table*}[htbp]
\centering
\caption{Ablation on the stitching ability}\label{tab2}
\begin{tabular}{|c|c|c|c|c|c|}
\hline\hline
\multicolumn{1}{|c}{}  & \multicolumn{1}{c}{\textbf{Dataset}} & \multicolumn{1}{c}{\textbf{CQL}} & \multicolumn{1}{c}{\textbf{DT}} & \multicolumn{1}{c}{\textbf{QT}} & \multicolumn{1}{c|}{\textbf{QDC}}  \\
\cline{1-6}
\hline
\multirow{4}{*}{\makecell{\textbf{Sparse}\\\textbf{Reward}}} 
& maze2d-open-v0 & 216.7 ± 80.7 & 196.4 ± 39.6 & 497.9 ± 12.3 & 479.3±4.6 \\
& maze2d-umaze-v1 & 94.7 ± 23.1 & 31.0 ± 21.3 & 105.4 ± 4.8 & \textbf{160.8}±9.1 \\
& maze2d-medium-v1 & 41.8 ± 13.6 & 8.2 ± 4.4 & 172.0 ± 6.2 & \textbf{172.8}±1.7 \\
& maze2d-large-v1 & 49.6 ± 8.4 & 2.3 ± 0.9 & 240.1 ± 2.5 & 238.5±7.3  \\
\hline
\multirow{4}{*}{\makecell{\textbf{Dense}\\\textbf{Reward}}}
& maze2d-open-dense-v0 & 307.6 ± 43.5 & 346.2 ± 14.3 & 608.4 ± 1.9 & 578.3±8.6 \\
& maze2d-umaze-dense-v1 & 72.7 ± 10.1 & -6.8 ± 10.9 & 103.1 ± 7.8 & 95.6±3.1 \\
& maze2d-medium-dense-v1 & 70.9 ± 9.2 & 31.5 ± 3.7 & 111.9 ± 1.9 & \textbf{131.9}±7.4 \\
& maze2d-large-dense-v1 & 90.9 ± 19.4 & 45.3 ± 11.2 & 177.2 ± 7.8 & 173.2±8.5 \\
\hline
\end{tabular}
\end{table*}

\section{Conclusion,Discussion and Future work}
In this paper, we propose a new decision-making algorithm for offline reinforcement learning (RL) called QDC. The DC component consists of convolutional filters combined with a Q-learning module for decision-making. Compared to the attention modules in DT, the convolutional module in DC reduces the number of parameters and computational complexity while better capturing local dependencies in RL offline trajectories. We have demonstrated that QDC performs exceptionally well across various offline RL environments, particularly in trajectory stitching. The convolutional structure of DC effectively captures local dependencies in RL trajectories, and the Q-learning module improves the selection of high-value actions. However, there are still some areas where performance could be improved. In future work, we will further optimize the structure of QDC and explore its potential in more practical applications.

\bibliographystyle{IEEETran}
\bibliography{mybib} 

\end{document}